\documentclass[letterpaper, 10pt]{article}
\usepackage{graphicx}
\usepackage{amsmath}
\usepackage{float}
\usepackage{amsfonts}
\usepackage[inner=0.75in, outer=0.75in, top=0.75in, bottom=1in, paperheight=11in, paperwidth=8.5in]{geometry}
\usepackage{algorithm}
\usepackage{algorithmicx}
\usepackage{algpseudocode}
\usepackage{epstopdf}
\usepackage{bm}
\usepackage{bbm}
\usepackage{tikz}
\usepackage{multirow}
\usepackage{amsmath}
\usetikzlibrary{bayesnet}
\usetikzlibrary{shapes.gates.logic.US,trees,positioning,arrows}
\usepackage{caption}
\usepackage{subcaption}
\usetikzlibrary{shapes,snakes}
\usetikzlibrary{trees}

\usepackage{amsthm}

\usepackage{epstopdf}
\usepackage{amssymb}
\usepackage{microtype}
\usepackage{url}
\usepackage{caption}
\usepackage{subcaption}

\usepackage[hidelinks]{hyperref}
\usepackage[capitalize]{cleveref}

\usepackage{xparse}

\usepackage[acronym,toc,nonumberlist]{glossaries}

\newacronym{SHM}{SHM}{Structural Health Monitoring}
\newacronym{PBSHM}{PBSHM}{Population-Based Structural Health Monitoring}
\newacronym{knn}{$k$-NN}{$k$-Nearest Neighbour}
\newacronym{tca}{TCA}{Transfer Component Analysis}
\newacronym{jda}{JDA}{Joint Domain Adaption}
\newacronym{artl}{ARTL}{Adaptation Regularization based Transfer Learning}
\newacronym{RKHS}{RKHS}{Reproducing Kernel Hilbert Space}
\newacronym{mmd}{MMD}{Maximum Mean Discrepancy}
\newacronym{pca}{PCA}{Principle Component Analysis}
\newacronym{SVM}{SVM}{Support Vector Machine}

\DeclareDocumentCommand{\pof}{m g}{ 
	{p( #1 %
		\IfNoValueF{#2}{\,\vert\, #2}%
		)%
	}
}

\DeclareDocumentCommand{\expect}{m g}{ 
	{\mathbb{E}\IfNoValueF{#2}{_{#2}}\left( #1 \right)%
	}
}

\DeclareDocumentCommand{\gaussianDist}{m m g}{ 
	{\mathcal{N}\left(\IfNoValueF{#3}{#3 \, \vert \,}
		#1,#2\right)%
	}
}

\begin{document}

	\title{\bf Towards risk-informed PBSHM: Populations as hierarchical systems}
	\author{A.J.\ Hughes, P.\ Gardner \& K.\ Worden \\ ~ \\
    Dynamics Research Group, Department of Mechanical Engineering, University of Sheffield, \\ Mappin Street, Sheffield S1 3JD, UK
	}
	\date{}
	\maketitle
	\thispagestyle{empty}
	
	\section*{Abstract}
	
	The prospect of informed and optimal decision-making regarding the operation and maintenance (O\&M) of structures provides impetus to the development of structural health monitoring (SHM) systems. A probabilistic risk-based framework for decision-making has already been proposed. The framework comprises four key sub-models: the utility model, the failure-modes model, the statistical classifier, and the transition model. The cost model consists of utility functions that specify the costs of actions and structural failures. The failure-modes model defines the failure modes of a structure as combinations of component and substructure failures via fault trees. The statistical classifier and transition model are models that predict the current and future health-states of a structure, respectively. Within the data-driven statistical pattern recognition (SPR) approach to SHM, these predictive models are determined using machine learning techniques. However, in order to learn these models, measured data from the structure of interest are required. Unfortunately, these data are seldom available across the range of environmental and operational conditions necessary to ensure good generalisation of the model.

	Recently, technologies have been developed that overcome this challenge, by extending SHM to \textit{populations} of structures, such that valuable knowledge may be transferred between instances of structures that are sufficiently similar. This new approach is termed population-based structural heath monitoring (PBSHM).

	The current paper presents a formal representation of populations of structures, such that risk-based decision processes may be specified within them. The population-based representation is an extension to the hierarchical representation of a structure used within the probabilistic risk-based decision framework to define fault trees. The result is a series, consisting of systems of systems ranging from the individual component level up to an inventory of heterogeneous populations. The current paper considers an inventory of wind farms as a motivating example and highlights the inferences and decisions that can be made within the hierarchical representation.

\textbf{Keywords: population-based structural health monitoring; risk; decision-making; value of information}
	
	\section{Introduction}
	
	Structural health monitoring (SHM) is a technology that aims to detect damage within  mechanical, civil and aerospace structures and infrastructure \cite{Farrar2013}. By inferring information about the health of a structure from discriminative features extracted from data acquired throughout a monitoring campaign, these systems can facilitate informed predictions relating to one or more of the following problems regarding the health of a structure, as summarised in Rytter's hierarchy \cite{Rytter1993}:

	\begin{itemize}
		\item The presence of damage in a structure (detection).
		\item The location of damage within a structure (localisation).
		\item The type of damage present in a structure (classification).
		\item The extent of damage in a structure (severity).
		\item The remaining safe/useful life of a structure (prognosis).
	\end{itemize}
	
	By informing predictions with data from a monitoring system, one can also inform decision-making regarding the operation and maintenance of structures and this can yield benefits such as improved safety, reduced operation costs and operational lifetime extension.

	Recent works have explicitly framed structural health monitoring in the context of decision-making \cite{Schobi2016,Vega2020a,Hughes2021}. The approach to decision-making for SHM presented in \cite{Hughes2021}, adopts a probabilistic risk-based perspective. In this approach, probability distributions over structural health states are inferred from data via statistical classifiers. The distributions are then forecast via a transition model and mapped to probabilities of failure for specific failure modes of interest via Bayesian network representations of fault trees. Optimal decisions are found by maximising the expected utility when considering both the risk of structural failure and the cost of maintenance actions. Several submodels have been identified as elements that are required to sufficiently define SHM decision processes; these submodels include statistical classifiers for inferring health-states and health-state transition models. In order to achieve robust decision-making, these submodels require labelled data for learning and/or validation.

	A critical challenge associated with the development of SHM systems is the scarcity of the data necessary for the learning and validation of models. Prior to the implementation of a monitoring system, there is often a lack of comprehensive labelled data across the health-states of interest for a given structure as obtaining data corresponding to damage states tends to  be prohibitively expensive or otherwise infeasible. One approach to circumvent this issue in the development of classification models is to utilise online active learning algorithms to preferentially obtain labelled data via structural inspections after a monitoring system is installed \cite{Bull2019,Hughes2022a,Hughes2022b}. In \cite{Vega2020b}, a methodology for determining a transition model using qualitative data from historical inspections is demonstrated.

	Population-based structural health monitoring (PBSHM), provides a holistic framework for overcoming data scarcity in the development of predictive models for SHM \cite{Bull2021,Gosliga2021,Gardner2021b,Tsialiamanis2021}. The core principal of PBSHM is that predictions about individual structures can be improved with the use of information transferred from other similar structures.

	The current paper aims to further the core principal of PBSHM, such that \emph{decisions} about the operation of both individual structures and populations of structures can be improved via the transfer of information. This is achieved by extending the hierarchical representation of structures, used to develop fault trees in the risk-based approach to decision-making for traditional SHM, to hierarchical representations of populations of structures. Throughout the current paper, an inventory of offshore wind farms are referenced as a motivating example.

	The layout of the current paper is as follows. Background theory is provided for both PBSHM and risk-based SHM in Sections \ref{sec:PBSHM} and \ref{sec:RBSHM}, respectively. Subsequently, a hierarchical representation of individual structures is presented in Section \ref{sec:Hierarchy_s}. This representation is extended to populations of structures in Section \ref{sec:Hierarchy_p}. Inferences and decisions within the population hierarchy are defined and discussed in Section \ref{sec:Inf_Dec}. Finally, conclusions are provided in Section \ref{sec:Conclusions}.

	\section{Population-based SHM}\label{sec:PBSHM}
	
	The foundations of PBSHM have been presented in a series of journal papers, each detailing the fundamental concepts of the approach; homogeneous populations \cite{Bull2021}, heterogeneous populations \cite{Gosliga2021}, mapping and transfer \cite{Gardner2021b}, and the geometric spaces in which structures exist \cite{Tsialiamanis2021}. By adopting a population-based approach to SHM, such that knowledge and information can be transferred between similar structures, there is the potential for improved diagnostic and prognostic capabilities \cite{Worden2020}.

	In the most general sense, a population can be considered to simply be a set of structures. Given the broad nature of this definition, in order to achieve useful transfer of knowledge and information between structures, it is discerning to consider specific classes of populations based upon the similarity of the constitutive structures. Thus, the notions of homogeneous and heterogeneous populations are introduced in \cite{Bull2021,Gosliga2021,Gardner2021b}.

	\subsection{Homogeneous and heterogeneous Populations}

	Within a population, structures may share common characteristics such as geometries, topologies, materials, and boundary conditions. Consider a population of wind turbines in an offshore wind farm and suppose these turbines are of the same model; developed to the same ISO standards and possessing common components, materials, aerodynamic design and so on. Qualitatively, these structures can be regarded as nominally identical. Populations comprised exclusively of nominally-identical structures are termed \textit{homogeneous populations}. Specific instances of structures in a homogeneous population can be considered to be perturbations of a population \textit{form} \cite{Bull2021}. For further discussions on population forms, the reader is directed to \cite{Bull2021}. Other examples of homogeneous populations include a fleet of Airbus A380s, an array of small modular nuclear reactors, and the Global Positioning System (GPS) satellite constellation.
	
	Variation between structures in homogeneous populations may arise because of factors such as environmental conditions and manufacturing defects. Returning to the example of an offshore wind farm, one could imagine that two turbines at differing locations in the farm may experience different geotechnical conditions -- perhaps as a result of varying geological composition in the seabed. Variability in such conditions could affect the boundary conditions of the monopile turbine towers and therefore modify the behaviours and data exhibited by these otherwise nominally-identical structures.

	In essence, \textit{heterogeneous} populations form the complement of the set of homogeneous populations \cite{Worden2020}; that is, heterogeneous populations are not exclusively comprised of structures that are nominally identical. Heterogeneous populations represent more general sets of structures and allow for differing designs, large variability in boundary conditions, and even multiple types of structure. While there may be stark differences between individual structures in a heterogeneous population, there may nonetheless be similarities that can be exploited to achieve useful knowledge and information transfer.
	
	Consider again the offshore wind farm example and suppose that the population is comprised of wind turbines each with three blades. Suppose also that the operating company manage an additional wind farm in a distinct location, comprised of four-blade turbines. Useful inferences could be achieved by considering these wind farms as two homogeneous populations, however, further insights could also be gained by considering them as a single heterogeneous population. For example, similarities may be present in the tower design between both types of wind turbine; hence, by considering a larger population from which to make observations, improved predictive models can be developed for this specific substructure. Other types of heterogeneous populations that may be useful to consider include inventories of aircraft comprised of a variety of models, and multiple suspension bridges with differing designs (e.g.\ single-span, multi-span).

	Thus far, similarities between structures have been described somewhat qualitatively, however, to better indicate where information transfer may work, it is useful to quantify this similarity.

	\subsection{Similarity between structures}

	Graph theory provides a rigorous and rich framework for representing and comparing discrete structured objects and has proved to be an invaluable modelling tool in fields such as chemistry and proteomics.

	In \cite{Gosliga2021}, the notion of the irreducible element (IE) model for structures is introduced as a representation of structures with relatively low-dimension when compared to alternatives such as finite element, or CAD models. The IE representation involves abstracting a structure into discrete elements having geometrical and material information (e.g.\ beams, plates, shells) and relationships (e.g.\ joints) so as to sufficiently capture the   nature of a structure. Here, the `nature' one wishes to capture pertains to health monitoring problems associated with a structure.

	Once an IE representation of a structure has been obtained, the information can be encoded into an attributed graph (AG). Whereas the purpose of the IE model is to present key characteristics of a structure in a human-readable format, the purpose of the AG is to embed a structure space, so as to facilitate the efficient pair-wise comparison of structures. With structures embedded into a metric space via AGs, one can utilise graph-matching algorithms to find common subgraphs between sets of structures. These subgraphs indicate substructures that are common within sets of structures and can be used to inform where transfer may be applicable. Furthermore, measures of closeness within the space of AGs (or common subgraphs) can be used to quantify similarity; in \cite{Gosliga2021} the Jaccard index is used and in \cite{Wickramarachchi2022} a variety of graph kernels are demonstrated.

	In summary, structures can be mapped into a graphical domain to facilitate comparison, identify common substructures and quantify similarity. By conducting this similarity assessment for structures within a population, one can determine where it is likely that information and knowledge can be successfully transferred between individual structures.

	\subsection{Mapping and Transfer}

	As mentioned previously, the primary benefit in taking a population-based approach to SHM is gaining the ability to transfer knowledge and information between sufficiently-similar individual structures; thereby overcoming issues associated with data scarcity.

	The sharing of knowledge and information between individual structures can be achieved via a number of methodologies. One manner in which this can be achieved is by having a statistical representation of the aforementioned population form, as demonstrated in \cite{Bull2021}. Another approach, presented in \cite{Dhada2020}, shares datasets in joint hierarchical statistical models of a population. Methodologies founded upon \textit{transfer learning} have also been successfully demonstrated \cite{Gardner2021b}. The principal of transfer learning is closely aligned with the goals of PBSHM; specifically, a branch of transfer learning known as \textit{domain adaptation}. In domain adaptation, datasets are adapted in a manner that allows a model constructed for a \textit{source} domain to generalise to a \textit{target} domain.

	For knowledge/information transfer to be successful, it is imperative that these source and target domains are comparable. This constraint can be adhered to by employing the similarity assessment outlined in the previous section.

	Thus far, PBSHM has been considered with respect to predictions and inferences. Before incorporating decisions into the PBSHM framework, background on the risk-based approach to decision-making for traditional SHM is provided.

	\section{Probabilistic risk-based SHM}\label{sec:RBSHM}

	The probabilistic risk-based approach to SHM is founded on the notion that monitoring systems should be designed and developed with consideration for the specific decision-support applications motivating their implementation.

	In the SHM paradigm detailed in \cite{Farrar2013}, monitoring campaigns begin with a process termed \textit{operational evaluation}. This stage in the development of an SHM system is concerned with specifying the context for an SHM system, dealing with aspects such as the safety/economic justification and the environmental and operational conditions. In \cite{Hughes2021}, it is proposed that the decision-making processes associated with an SHM campaign should be considered from the outset of a monitoring campaign as part of the operational evaluation. 
	
	To begin defining the decision processes that one may wish to inform with a monitoring system, one must identify a set of failure modes or conditions for a structure that one may wish to prevent, in addition to a set of actions that can be executed to aid in the mitigation of failures. Furthermore, as part of the economic justification of a monitoring system, costs or utilities must be assigned to these failures and actions. The prediction of specific failure events and the informed selection of optimal mitigating actions should provide the basis for the development of monitoring systems guiding, choices for aspects of the monitoring system such as: sensors and their placement on the structure; data processing; and the discriminative features and models used to classify structural health states.

	Once a monitoring system developed with respect to decision-making is implemented, optimal strategies can be found by maximising expected utility with consideration for the risk of failure and the cost of mitigating actions. This can be achieved by representing the decision processes as a probabilistic graphical model (PGM). The following two subsections provide background on PGMs and the modelling of SHM decision processes as PGMs respectively.

	\subsection{Probabilistic graphical models}

	Probabilistic graphical models (PGMs) are graphical representations of factorisations of joint probability distributions, and are a powerful tool for reasoning and decision-making under uncertainty. For this reason, they are apt for representing and solving decision problems in the context of SHM, where there is uncertainty in the health states of structures. While there exist multiple forms of probabilistic graphical model, the key types utilised for representing SHM decision processes are Bayesian networks (BNs) and influence diagrams (IDs) \cite{Sucar2015}.

Bayesian networks are directed acyclic graphs (DAGs) comprised of nodes and edges. Nodes represent random variables, and edges connecting nodes represent conditional dependencies between variables. In the case where the random variables in a BN are discrete, the model is defined by a set of conditional probability tables (CPTs). For continuous random variables, the model is defined by a set of conditional probability density functions (CPDFs).

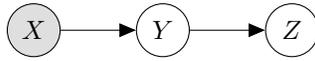
\begin{figure}[ht!]
	\centering
	\begin{tikzpicture}[x=1.7cm,y=1.8cm]
	

	\node[obs] (parent) {$X$} ;
	\node[latent, right=1cm of parent] (inter) {$Y$} ;
	\node[latent, right=1cm of inter] (child) {$Z$} ;
	
	\edge {parent} {inter} ; %
	\edge {inter} {child} ; %
	
	\end{tikzpicture}
	\caption{An example Bayesian network.}
	\label{fig:PGM0}
\end{figure}

Figure \ref{fig:PGM0} shows a simple Bayesian network comprised of three random variables $X$, $Y$ and $Z$. $Y$ is conditionally dependent on $X$ and is said to be a \textit{child} of $X$, while $X$ is said to be a \textit{parent} of $Y$. $Z$ is conditionally dependent on $Y$ and can be said to be a child of $Y$ and a \textit{descendant} of $X$, while $X$ is said to be an \textit{ancestor} of $Z$. The factorisation described by the Bayesian network shown in Figure \ref{fig:PGM0} is given by $P(X,Y,Z) = P(X)\cdot P(Y|X)\cdot P(Z|Y)$. Given observations on a subset of nodes in a BN, inference algorithms can be applied to compute posterior distributions over the remaining unobserved variables. Observations of random variables are denoted in a BN via grey shading of the corresponding nodes, as is demonstrated for $X$ in Figure \ref{fig:PGM0}.

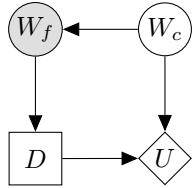
\begin{figure}[ht!]
	\centering
	\begin{tikzpicture}[x=1.7cm,y=1.8cm]
	

	\node[latent] (condition) {$W_c$} ;
	\node[obs,left=1cm of condition] (forecast) {$W_f$} ;
	\node[rectangle,draw=black,minimum width=0.7cm,minimum height=0.7cm,below=1cm of forecast] (decision) {$D$} ;
	\node[det, below=1cm of condition] (utility) {$U$} ;
	
	\edge {condition} {forecast} ; %
	\edge {forecast} {decision} ; %
	\edge {condition} {utility} ; %
	\edge {decision} {utility} ; %
	
	\end{tikzpicture}
	\caption{An example influence diagram representing the decision of whether to go outside or stay in under uncertainty in the future weather condition given an observed forecast.}
	\label{fig:ID0}
\end{figure}

Bayesian networks may be adapted into influence diagrams to model decision problems. This augmentation involves the introduction of two additional types of node, as shown in Figure \ref{fig:ID0}: decision nodes, denoted as squares, and utility nodes, denoted as rhombi. For influence diagrams, edges connecting random variables to utility nodes denote that the utility function is dependent on the states of the random variables. Similarly, edges connecting decisions nodes to utility nodes denote that the utility function is dependent on the decided actions. Edges from decision nodes to random variable nodes indicate that the random variables are conditionally dependent on the decided actions. Edges from random variable or decision nodes to other decision nodes do not imply a functional dependence but rather order, i.e.\ that the observations/decisions must be made prior to the next decision being made.

To gain further understanding of IDs, one can consider Figure \ref{fig:ID0}. Figure \ref{fig:ID0} shows the ID for a simple binary decision; stay home and watch TV or go out for a walk, i.e.\ $\text{dom}(D) = \{ \text{TV},\text{ walk} \} $. Here, the agent tasked with making the decision has access to a weather forecast $W_{f}$ which is conditionally dependent on the future weather condition $W_{c}$. The weather forecast and future condition share the same possible states $\text{dom}(W_f) = \text{dom}(W_c) = \{ \text{bad}, \text{ good} \} $. The utility achieved $U$, is then dependent on both the future weather condition and the decided action. For example, one might expect high utility gain if the agent decides to go for a walk and the weather condition is good.

In general, a policy $\delta$ is a mapping from all possible observations to possible actions. The problem of inference in influence diagrams is to determine an optimal strategy $\bm{\Delta}^{\ast} = \{ \delta^{\ast}_{1},\ldots, \delta^{\ast}_{n} \}$ given a set of observations on random variables, where $\delta^{\ast}_{i}$ is the $i^{th}$ decision to be made in a strategy $\bm{\Delta}^{\ast}$ that yields the \textit{maximum expected utility} (MEU). For further details on the computation of the MEU for influence diagrams, the reader is directed to \cite{Kjaerulff2008}. Defined as a product of probability and utility, the expected utility can be considered as a quantity corresponding to risk.

	\subsection{Decision framework}

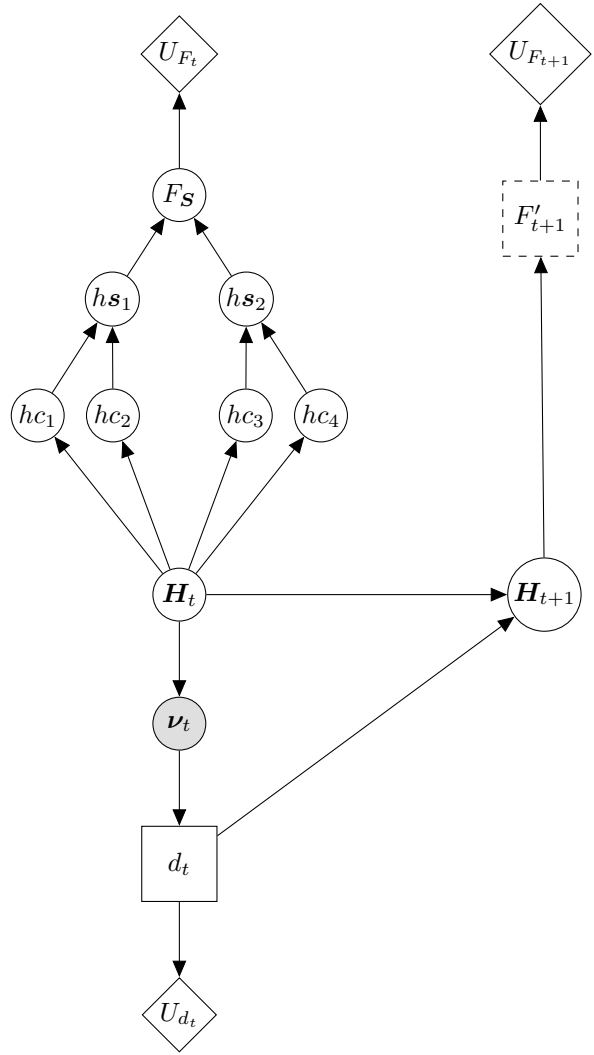
\begin{figure}[h!]
	\centering
	\begin{tikzpicture}[x=1.7cm,y=1.8cm]
	
	\node[det] (uf1) {$U_{F_{t}}$} ;
	\node[latent,below=1cm of uf1] (failure) {$F_{\bm{S}}$} ;
	\node[const, below=1cm of failure] (temp) {$ $} ;
	\node[latent, right=0.5cm of temp] (bay2) {$h\bm{s}_{2}$} ;
	\node[latent, left=0.5cm of temp] (bay1) {$h\bm{s}_{1}$} ;
	\node[const, below=1.5cm of temp] (temp2) {$ $} ;
	\node[latent, right=0.5cm of temp2] (mem6) {$hc_{3}$} ;
	\node[latent, right=1.5cm of temp2] (mem2) {$hc_{4}$} ;
	\node[latent, left=1.5cm of temp2] (mem5) {$hc_{1}$} ;
	\node[latent, left=0.5cm of temp2] (mem1) {$hc_{2}$} ;
	\node[det, right=3.6cm of uf1] (uf2) {$U_{F_{t+1}}$} ;
	\node[rectangle,draw=black,dashed,minimum width=1cm,minimum height=1cm,below=1cm of uf2] (ft2) {$F^\prime_{t+1}$} ;
	\node[latent, below=2cm of temp2] (x1) {$\bm{H}_{t}$} ;
	\node[latent, right=4cm of x1] (x2) {$\bm{H}_{t+1}$} ;
	\node[obs, below=1cm of x1] (y1) {$\bm{\nu}_{t}$} ;
	\node[rectangle,draw=black,minimum width=1cm,minimum height=1cm,below=1cm of y1] (d1) {$d_{t}$} ;
	\node[det, below=1cm of d1] (u1) {$U_{d_{t}}$} ;
	
	\edge {failure} {uf1} ; %
	\edge {ft2} {uf2} ; %
	\edge {x2} {ft2} ; %
	\edge {x1} {x2} ; %
	\edge {x1} {y1} ; %
	\edge {d1} {x2} ; %
	\edge {d1} {u1} ; %
	\edge {bay1} {failure} ; %
	\edge {bay2} {failure} ; %
	\edge {mem1} {bay1} ; %
	\edge {mem5} {bay1} ; %
	\edge {mem2} {bay2} ; %
	\edge {mem6} {bay2} ; %
	\edge {x1} {mem1} ; %
	\edge {x1} {mem5} ; %
	\edge {x1} {mem2} ; %
	\edge {x1} {mem6} ; %
	\edge {y1} {d1} ; %

	\end{tikzpicture}
	\caption{An influence diagram representing a partially-observable Markov decision process over one time-slice for determining the utility-optimal maintenance strategy for a simple structure comprised of four components. The fault-tree failure-mode model for time $t+1$ has been represented as the node $F^\prime_{t+1}$ for compactness.}
	\label{fig:OverallPGM1}
\end{figure}

	A probabilistic graphical model for a general SHM decision problem across a single time-slice is shown in Figure \ref{fig:OverallPGM1}. Here, a maintenance decision $d$ is shown for a simple fictitious structure $\bm{S}$, comprised of two substructures $\bm{s}_{1}$ and $\bm{s}_{2}$, each of which are comprised of two components; $c_{1,2}$ and $c_{3,4}$, respectively.

The overall decision process model shown in Figure \ref{fig:OverallPGM1} is based upon a combination of three sub-models; a statistical classifier, a failure-mode model, and a transition model.

The failure condition of the structure $F_{\bm{S}}$ is represented as a random variable within the PGM. This failure condition $F_{\bm{S}}$ can be expressed as a failure mode of the global structure that can be specified by a fault tree; a combination of local failures at a component, joint and substructure level related by Boolean logic gates. In order to fit into the PGM framework, fault trees must be mapped into Bayesian networks. Fortunately, there is a well-established mapping for this presented in \cite{Bobbio2001,Mahadevan2001}. Essentially, components, joints and substructures receive random variables in the PGM corresponding to their respective local health states. The conditional probability tables defining the relationship between these random variables correspond to the Boolean truth tables for each of the logic gates in the fault tree defining the failure mode $F_{\bm{S}}$.

Figure \ref{fig:OverallPGM1} considers a failure mode dependent on two substructures $\bm{s}_{1-2}$, which in turn are each dependent on two components $c_{1,2}$ and $c_{3,4}$.  For the decision process shown in Figure \ref{fig:OverallPGM1}, component health states are denoted by $hc_{1-4}$, and substructure health states are denoted by $h\bm{s}_{1,2}$.  An advantage of considering specific failure modes and their representations as fault trees is that doing so yields the health states that must be targeted by the monitoring system; the local health states of the components can be summarised in a global health-state vector. For the example shown in Figure \ref{fig:OverallPGM1}, this health-state vector is given by $\bm{H} = \{ hc_1, hc_2, hc_3, hc_4 \}$. In essence, the purpose of this failure model is to map from a distribution over the global health states to a probability of structural failure. Finally, the failure states associated with the variable $F_{\bm{S}}$ are given utilities via the function represented by the node $U_F$. As it is necessary to consider the future risk of failure in the decision process, this failure-mode model and utility function are repeated for each time-step.

As previously mentioned, a random variable denoted $\bm{H}_t$ is used to represent the latent global health state of the structure at time $t$. Within the decision process, the function of the statistical classifier is to provide a posterior probability distribution over the latent health state $\bm{H}_t$, inferred via observations on a set of discriminative features $\bm{\nu}_t$. This probability distribution over health-states may be obtained via a generative model $P(\bm{\nu}|\bm{H})$ as shown in Figure \ref{fig:OverallPGM1}, or obtained more directly via a discriminative classifier which yields $P(\bm{H}|\bm{\nu})$. Here, the use of a probabilistic classifier is vital to ensure decisions made are robust to uncertainty in the health state of the structure.

Finally, a transition model is used to forecast the future health states, given the current health state and a decided action, i.e.\ $P(\bm{H}_{t+1}|\bm{H}_{t},d_{t})$. The transition model considers the degradation of the structure under the various operational and environmental conditions a structure may experience, while accounting for uncertainties in each.

By employing decision-process models such as the one presented here, one can obtain optimal strategies regarding the operation and maintenance of individual structures by maximising the expected utility.
	
	\section{Structures as hierarchies}\label{sec:Hierarchy_s}

	A key assumption implicit in the development of the fault-tree failure models within the risk-based SHM decision framework, is that structures can be represented as a hierarchy, or, in other terms, as a system of systems of systems. As it is outside the scope of the current paper, a comprehensive and consistent notation for referencing specific elements of a structure is not established here. Rather, the constituent levels and elements within the hierarchical representation are presented, in addition to the process by which one arrives at them. 

	Consider a structure of interest $S$. To obtain a hierarchical representation for $S$, one must first decompose $S$ into a discrete number of constituent elements, which are referred to as \textit{substructures}. Substructures are considered to be entities which may, in principle, be assembled remotely or available for independent testing prior to incorporation into the full-scale structure. Within the hierarchical representation, some substructures may be further decomposed up until the stage at which it would no longer be meaningful or useful to do so. Substructures at this stage are referred to as \textit{components}. As such, components are considered to be substructures which cannot (or need not) be decomposed further; these are the smallest element of a structure one might reasonably monitor. A notable sub-class of component is the \textit{joint}. Joints are considered to be the physical mechanisms by which substructures are joined together.

\begin{figure}[ht!]
		\centering
		\begin{tikzpicture}

	\node[const] (dots1) { } ;
	\node[rectangle, below=0.2cm of dots1, draw=black,minimum width=15cm,minimum height=1.5cm] (S3) {$\mathcal{S}^{3}:$ Structure, $S$} ;
	\node[const,below=1cm of S3] (temp1) { } ; 
	\node[rectangle, left=2cm of temp1, draw=black,minimum width=5.5cm,minimum height=1.5cm] (S21) {$\mathcal{S}^{2}$: Substructure, $s_{1}$} ;
	\node[rectangle, left=-2.3cm of temp1, draw=black,minimum width=4cm,minimum height=1.5cm] (S22) {$s_{2}$} ;
	\node[rectangle, left=-6.6cm of temp1, draw=black,minimum width=4cm,minimum height=1.5cm] (S23) {$s_{3}$} ;
	\node[const,right=6.9cm of temp1] (sCont) {$\cdots$} ;
	\node[const, below=1.75cm of temp1] (temp2) { } ; 
	\node[rectangle, left=3cm of temp2,, draw=black,minimum width=4.5cm, minimum height=1.5cm, text width=3.8cm,align=center] (S11) {$\mathcal{S}^{1}$: Component, $c_{1}$} ;
	\node[const,right=0.25cm of S11] (cCont1) {$\cdots$} ;
	\node[rectangle, left=-1.5cm of temp2, draw=black,minimum width=1.5cm,minimum height=1.5cm] (S13) {$c_{2}$} ;
	\node[rectangle, left=0.2cm of temp2, draw=black,minimum width=1.5cm,minimum height=1.5cm] (S12) {$c_{1}$} ;
	\node[const,right=0.1cm of S13] (cCont2) {$\cdots$} ;
	\node[rectangle, left=-5.8cm of temp2, draw=black,minimum width=1.5cm,minimum height=1.5cm] (S15) {$c_{2}$} ;
	\node[rectangle, left=-4.1cm of temp2, draw=black,minimum width=1.5cm,minimum height=1.5cm] (S14) {$c_{1}$} ;
	\node[const,right=0.1cm of S15] (cCont3) {$\cdots$} ;
	\node[const,right=0.5cm of cCont3] (cCont4) {$\cdots$} ;
\end{tikzpicture}
		\caption{A structure as systems of systems.}
		\label{fig:systemscont}
	\end{figure}
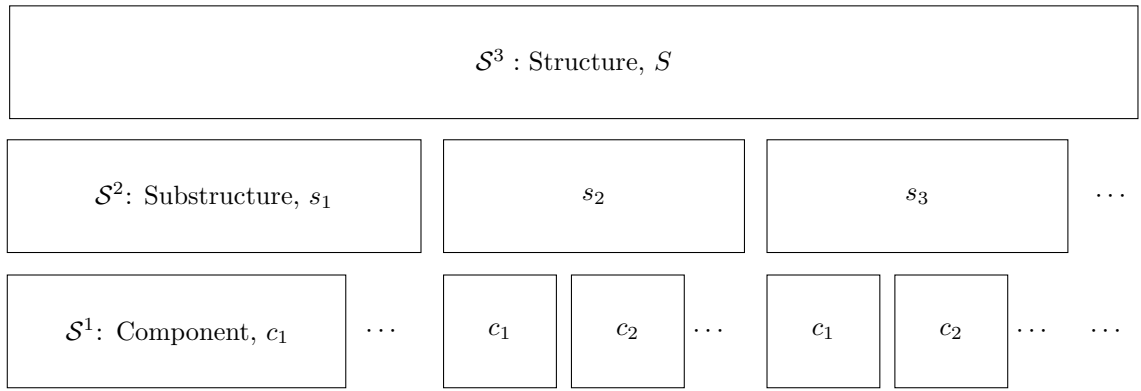

	A diagram illustrating the hierarchical representation of a structure is shown in Figure \ref{fig:systemscont}. The levels in the hierarchy that specifies the system of systems of systems shown are denoted as $\mathcal{S}^1$, $\mathcal{S}^2$, and $\mathcal{S}^3$ -- corresponding to the component, substructure and substructure levels, respectively. Within each level of the hierarchy, elements can be listed. 
	
	Returning to the example of a wind farm, it would be perfectly reasonable to consider a single turbine as an individual structure, representing the $\mathcal{S}^3$ level in a hierarchy. In the $\mathcal{S}^2$ level one may consider substructures such as the drive train, blades, or tower. Finally, in the $\mathcal{S}^1$ level one may have components such as the gearbox or bearings comprising the drive train, or the web and shells comprising the blades.

	The hierarchical representation of structures facilitates the specification of the decision process that motivate the development and implementation of SHM technologies. This facilitation is achieved by decomposing structures into constituent substructures and components which can then be used to define failure modes of the structure. Given a finite set of failure modes of interest, one can then specify critical components, and therefore health states, to be targeted by a monitoring system. 

	\section{Populations of structures as hierarchies}\label{sec:Hierarchy_p}

	A natural method for incorporating decision-making into PBSHM, is to extend the hierarchical representation of structures to hierarchical representations of populations. The number of levels required in a hierarchy is of course dependent on context. However, it is deemed that an additional three levels provide sufficient  generality for most PBSHM applications, and indeed the discussions in the current paper.

	The additional levels necessary to extend the hierarchical representation to populations of structures can be summarised as follows:

	\begin{itemize}
		\item $\mathcal{S}^{4}$ -- Type/Model Inventory: This level of the hierarchy corresponds to the lowest population level and represents an organisational grouping in which all individual structures in the population are of the same type/model and can be considered to be nominally identical. Thus, populations at this level in the hierarchy are homogeneous.
		\item  $\mathcal{S}^{5}$ -- Group Inventory: This next population level corresponds to a set of $\mathcal{S}^4$ inventories for which it is necessary or convenient to consider as a group for operational reasons such as asset management. As a group inventory may be formed of disparate type/model inventories, in general, group inventories are heterogeneous populations.
		\item $\mathcal{S}^6$ -- Inventory: This level of the hierarchy corresponds to the total set of structural assets operated or owned by an organisation or company. Again, this level will generally represent a heterogeneous population.
	\end{itemize}

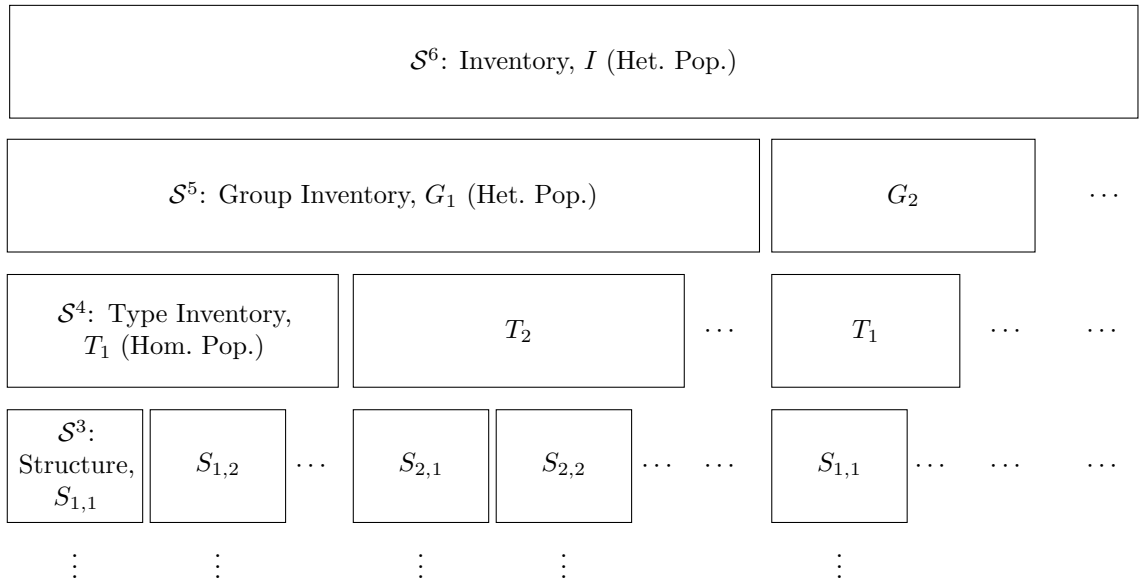
\begin{figure}[ht!]
		\centering
		\begin{tikzpicture}
	
	\node[rectangle,draw=black,minimum width=15cm,minimum height=1.5cm] (S6) {$\mathcal{S}^{6}$: Inventory, $I$ (Het.\ Pop.)} ;
	\node[const, below=1cm of S6] (temp1) { } ; 
	\node[rectangle, left=-2.5cm of temp1, draw=black,minimum width=10cm,minimum height=1.5cm] (S51) {$\mathcal{S}^{5}$: Group Inventory, $G_1$ (Het.\ Pop.)} ;
	\node[rectangle, right=2.6cm of temp1, draw=black,minimum width=3.5cm,minimum height=1.5cm] (S52) {$G_2$} ;
	\node[const, right=0.7cm of S52] (Gcont) {$\cdots$} ;
	\node[const, below=1.75cm of temp1] (temp2) { } ; 
	\node[rectangle, left=-1.5cm of temp2, draw=black,minimum width=4.4cm,minimum height=1.5cm] (S42) {$T_2$} ;
	\node[rectangle, left=3.1cm of temp2, draw=black,minimum width=4.4cm,minimum height=1.5cm,text width=4cm,align=center] (S41) {$\mathcal{S}^{4}$: Type Inventory, $T_1$ (Hom.\ Pop.)} ;
	\node[const, right=1.7cm of temp2] (Tcont1) {$\cdots$} ;
	\node[rectangle, right=2.6cm of temp2, draw=black,minimum width=2.5cm,minimum height=1.5cm] (S43) {$T_1$} ;
	\node[const, right=5.5cm of temp2] (Tcont2) {$\cdots$} ;
	\node[const, right=0.8cm of Tcont2] (Tcont3) {$\cdots$} ;
	\node[const, below=1.75 of temp2] (temp3) { } ; 
	\node[rectangle, left=3.8cm of temp3,draw=black,minimum width=1.8cm,minimum height=1.5cm] (S32) { $S_{1,2}$} ;
	\node[rectangle, left=5.7cm of temp3,draw=black,minimum width=1.8cm,minimum height=1.5cm,text width=1.5cm,align=center] (S31) {$\mathcal{S}^{3}$: Structure, $S_{1,1}$} ;
	\node[const, left=3.2cm of temp3] (Scont1) {$\cdots$} ;
	\node[rectangle, left=-0.8cm of temp3,draw=black,minimum width=1.8cm,minimum height=1.5cm] (S34) { $S_{2,2}$} ;
	\node[rectangle, left=1.1cm of temp3,draw=black,minimum width=1.8cm,minimum height=1.5cm] (S33) { $S_{2,1}$} ;
	\node[const, left=-1.4cm of temp3] (Scont2) {$\cdots$} ;
	\node[const, right=1.7cm of temp3] (Scont3) {$\cdots$} ;
	\node[rectangle, right=2.6cm of temp3,draw=black,minimum width=1.8cm,minimum height=1.5cm] (S35) { $S_{1,1}$} ;
	\node[const, right=4.5cm of temp3] (Scont4) {$\cdots$} ;
	\node[const, right=5.5cm of temp3] (Scont5) {$\cdots$} ;
	\node[const, right=0.8cm of Scont5] (Scont6) {$\cdots$} ;
	\node[const, below=0.2cm of S31] (Ccont1) {$\vdots$} ;
	\node[const, below=0.2cm of S32] (Ccont2) {$\vdots$} ;
	\node[const, below=0.2cm of S33] (Ccont3) {$\vdots$} ;
	\node[const, below=0.2cm of S34] (Ccont4) {$\vdots$} ;
	\node[const, below=0.2cm of S35] (Ccont5) {$\vdots$} ;
	
\end{tikzpicture}
		\caption{An inventory as a system of systems of systems of systems.}
		\label{fig:systems}
	\end{figure}

Figure \ref{fig:systems} depicts the continuation of the hierarchical representation from $\mathcal{S}^3$ to $\mathcal{S}^6$. In Figure \ref{fig:systems}, an inventory $I$ is considered as a system of systems of systems of systems. Once again, a list can be formed of the constituent elements for each level in the hierarchy.

To further elucidate this extension of the hierarchy, once again, consider the example of an organisation operating offshore wind farms. As previously indicated, a wind farm comprised exclusively of turbines of a single type or model can form a homogeneous population; this corresponds to $\mathcal{S}^4$ in the hierarchy. In the case that the organisation is responsible for multiple wind farms, or a single farm with a mixture of turbine types, one may wish to organise these type/model inventories into group inventories. For example, these group inventories may be formed from type inventories according to the geographical jurisdiction of sub-divisions within the organisation, or even formed from a collection of type inventories that are overseen by a single maintenance crew. Should these populations each be comprised of a different model of wind turbine, the group inventories formed would be heterogeneous populations and correspond to $\mathcal{S}^5$ in the hierarchy. Alternatively, if all the wind farms consist of a single type of turbine, $\mathcal{S}^4$ and $\mathcal{S}^5$ can be merged and the group inventories are instead homogeneous populations. Finally, the group inventories owned by the wind farm organisation can be aggregated as an inventory in the $\mathcal{S}^6$ level of the hierarchy. This level would represent the organisation's total structural assets and could amount to, for example, multiple wind farms spread across the globe, maritime vessels, and aircraft that may be used for inspection, maintenance or other operational activities.

As is the case for traditional SHM, the hierarchical representation of structures and populations of structures can help facilitate decision-making for PBSHM in several ways. These decision processes are discussed further in the following section.

	\section{Risk-informed PBSHM}\label{sec:Inf_Dec}

	Numerous decisions must be made throughout the life cycle of a PBSHM system. Most obvious are the operation and maintenance decisions an organisation may have to make, following the installation of a monitoring system, such as inspections and repairs. Equally important, however, are the decisions that must be made prior to implementation such as those made in the operational evaluation stage of PBSHM.

	\subsection{Operational evaluation}
	
	One significant way in which adopting a hierarchical risk-based approach to PBSHM facilitates decision-making occurs very early on, in the operational evaluation stage. By considering specific failure modes and constructing fault trees for individual structures, one can decide the key elements of a structure which should be modelled in IEs and AGs. In other words, the specification of failure modes as combinations of component and substructure failures can be used to inform the granularity at which IEs and AGs are constructed. A further benefit of the population-based approach is that, as structures are considered nominally identical, large proportions of the fault trees may be mapped across a homogeneous population, with the exception of perhaps environment-specific failure modes.
	
	The extension of the hierarchy to represent populations of structures via the inclusion of levels $\mathcal{S}^4$ to $\mathcal{S}^6$ prompts one to consider how failures may be defined at the population level. One possible way to approach the failure of a population would be to consider the critical missions for the operating organisation. Depending on the nature of the organisation -- whether they are non-commercial or commercial -- these missions may be related to performance measures such as availability and/or profitability. Consider the wind farm example. Suppose that the operating organisation are required to supply energy from the wind farm to an electrical grid while maintaining a total population availability of 99\%. This population can then be considered to have failed if the population structural availability falls below 99\%. This population failure may be specified then by extending the fault tree; defining the population failures as a combination of individual failures. In addition, the organisation may wish to specify a failure condition based upon profitability, perhaps based upon a performance criterion related to a moving-average of the total power output. Again, this failure could be represented as a combination of individual structure failures and environmental conditions. This distinct failure mode is likely to be highly correlated with the availability failure mode; fortunately, the probabilistic graphical models employed in the risk-based approach can account for these `common-cause' failures. This approach to defining population failures can be applied at any of the population levels within the hierarchical representation by considering combinations of failures in the levels below.

	Defining failures at the population level within the hierarchy allows one to assign costs during the operational evaluation stage. Following on from this, population-scale actions can be also be defined. 

	\subsection{Inferences and decisions}

	A fundamental process of decision-making for PBSHM is reasoning under uncertainty. This is typically achieved via inferences. Within the hierarchical framework for PBSHM, different types of inferences can be defined:

	\begin{itemize}
		\item I-inference: This type of inference corresponds to those usually made in traditional SHM, and occur within the individual structure levels $\mathcal{S}^3$ to $\mathcal{S}^1$. An example of an I-inference is the process of determining a probability distribution over the health states of an individual structure using data acquired from that structure.
		\item L-inference: This type of inference occurs between levels in the hierarchical representation of structures. These may also be types of I-inference, for example determining the probability of failure for a (sub)structure given local component health states. Other L-inferences may include those relating to the validation and verification of predictive models (V\&V). For example, one may be able to validate a predictive model for a structure at the $\mathcal{S}^3$ level with data measured from substructures or components at the $\mathcal{S}^2$ and $\mathcal{S}^1$ levels, respectively.
		\item P-inference: This type of inference occurs across populations. If the inference is across a type inventory in $\mathcal{S}^4$, i.e.\ a homogeneous population, they can be denoted as HomP-inferences. These inferences across populations may utilise technologies such as forms \cite{Bull2021}. An example of a HomP-inference is inferring the health state of a member in a population using data aggregated across all members in the population. On the other hand, if a P-inference is between populations containing different types of structure, such as within a group inventory in $\mathcal{S}^5$, then the inferences can be referred to as HetP-inferences. HetP-inferences may involve using transfer learning techniques such as domain adaptation \cite{Gardner2021b}. An example of a HetP-inference is transferring the degradation (transition) model for a blade from a population of four-blade wind turbines to a population of three-blade wind turbines.
	\end{itemize}

These inferences within the hierarchical representation of populations, facilitate reasoning under uncertainty using PBSHM systems; this can naturally be extended to decision-making under uncertainty, by considering the following types of decision:
	
	\begin{itemize}
		\item I-decision: This type of decision is made at the individual structure levels in the hierarchy, $\mathcal{S}^1$ to $\mathcal{S}^3$. Again, this type of decision corresponds to decisions one may make with a traditional SHM system. An example of an I-decision is selecting a maintenance strategy for an individual structure, substructure, or component for repair. Unlike in traditional SHM, in the risk-informed PBSHM approach, I-decisions can be informed by I-, L- and P-inferences alike.
		\item L-decision: The actions selected via this type of decision operate between levels of the hierarchical representation. As with L-inferences these decisions may pertain to the V\&V of predictive models. For example, deciding whether can one proceed with using a structural model validated on substructures. Another example of this type of decision relates to resource allocation. Suppose one has a limited budget to carry out some structural testing to acquire data for mode updating. Under these circumstances, one should aim to decide on a set of tests, and the levels at which these tests are carried out, such that the largest improvement in model performance is obtained for the given budget.
		\item P-decision: This type of decision is made at the population levels in the hierarchy, $S^4$ to $S^6$. These actions may pertain to resource management. For example, one may decide to send a team of engineers to perform inspections on a type inventory based on the probability of failure for a population rather than the probability of failure of an individual structure. Scheduling inspections in this manner could save both time and expenditure. Again, these decisions may be informed via I-, L- and P-inferences.
	\end{itemize}

	To summarise, the hierarchical representation of populations of structures facilitates both making inferences and making decisions for PBSHM, by allowing for the definition of specific types of inferences and decisions.

	\subsection{Value of information transfer}\label{sec:VOIT}
	
	Value of information (VoI) is a concept in decision theory defined to be the amount of money/resource a decision-maker should be willing to pay in order to gain access to information prior to making a decision. The concept of VoI has seen some application to traditional SHM in recent works \cite{Hughes2021b,Kamariotis2022}.
	
	Extending the risk-based approach to decision-making from traditional SHM to PBSHM opens up the possibility of value of information transfer, i.e.\ the price a decision-maker should be willing to pay in order to gain information via transfer, prior to making a decision. This value arises as a result of change in maximum expected utility that can be achieved should a change in optimal policy occur as a result of the additional information made available via transfer. This notion of value of information transfer yields the thought-provoking implication that, in some contexts, it may be an optimal decision to allow a (sub)structure to fail, since the data obtained throughout the failure process may improve the management of the other individuals in a population.

	\section{Conclusions}\label{sec:Conclusions}
	
	To conclude, PBSHM provides a general framework for overcoming issues of data scarcity associated with developing predictive models for detecting and forecasting damage within structures. This advantage is achieved via technologies that allow for the transfer of information between individual structures within a population. Adopting a probabilistic risk-based approach to SHM, allows inferences made about the health-states of individual structures to inform operation and maintenance decisions via the use of hierarchical representations of structures and fault trees. The current paper extends this hierarchical representation of structures to representations of populations, such that decision process can be defined over populations. Other advantages can be gained by adopting a risk-based approach to PBSHM; for example, the identification of critical components and substructures can be used to inform the development of irreducible element models and the associated attributed graphs. 
	
	\section*{Acknowledgements}
	The authors would like to gratefully acknowledge the support of the UK Engineering and Physical Sciences Research Council (EPSRC) via grant references EP/W005816/1 and EP/R006768/1. For the purpose of open access, the authors has applied a Creative Commons Attribution (CC BY) licence to any Author Accepted Manuscript version arising. KW would also like to acknowledge support via the EPSRC Established Career Fellowship EP/R003625/1.
	
	\bibliographystyle{unsrt}
	\bibliography{IMAC2022}

\end{document}